\documentclass{article}

\usepackage{PRIMEarxiv}

\usepackage{cite}
\usepackage{amsmath,amssymb,amsfonts}
\usepackage{graphicx}
\usepackage[table,xcdraw]{xcolor}
\usepackage{colortbl}
\usepackage{textcomp}
\usepackage{multirow}
\usepackage{pdfpages}
\usepackage{subcaption}

\usepackage[utf8]{inputenc} 
\usepackage[T1]{fontenc}    
\usepackage{hyperref}       
\usepackage{url}            
\usepackage{booktabs}       
\usepackage{amsfonts}       
\usepackage{nicefrac}       
\usepackage{microtype}      
\usepackage{lipsum}
\usepackage{fancyhdr}       
\usepackage{graphicx}       
\graphicspath{{media/}}     
\usepackage{algorithm, algpseudocode}

\def\BibTeX{{\rm B\kern-.05em{\sc i\kern-.025em b}\kern-.08em
    T\kern-.1667em\lower.7ex\hbox{E}\kern-.125emX}}

\algnewcommand{\Inputs}[1]{%
  \Statex \textbf{Inputs:}
  \Statex \hspace*{\algorithmicindent}\parbox[t]{.8\linewidth}{\raggedright #1}
}
\algnewcommand{\Outputs}[1]{%
  \Statex \textbf{Outputs:}
  \Statex \hspace*{\algorithmicindent}\parbox[t]{.8\linewidth}{\raggedright #1}
}
\algnewcommand{\StepOne}[1]{%
  \Statex \textbf{Step One:}
}
\algnewcommand{\StepTwo}[1]{%
  \Statex \textbf{Step Two:}
}

\title{Soft Segmented Randomization: Enhancing Domain Generalization in SAR-ATR for Synthetic-to-Measured}

\author{
    Minjun Kim$^1$, Ohtae Jang$^2$, Haekang Song$^1$, Heesub Shin$^3$, Jaewoo Ok$^3$,\\ \textbf{Minyoung Back$^3$, Jaehyuk Youn$^3$, and Sungho Kim$^1$}\\ \\
    $^1$Department of Electronic Engineering, Yeungnam University, South Korea\\
    $^2$Department of Electrical Engineering, POSTECH, South Korea\\
    $^3$LIG Nex1 Co., Ltd., South Korea\\ 
    \texttt{sunghokim@yu.ac.kr}
}

\begin{document}
\maketitle

\begin{abstract}
Synthetic aperture radar technology is crucial for high-resolution imaging under various conditions; however, the acquisition of real-world synthetic aperture radar data for deep learning-based automatic target recognition remains challenging due to high costs and data availability issues. To overcome these challenges, synthetic data generated through simulations have been employed, although discrepancies between synthetic and real data—stemming from factors such as background clutter and target signature differences—can degrade model performance. In this study, we introduce a novel framework, soft segmented randomization, designed to reduce domain discrepancy and improve the generalize ability of synthetic aperture radar automatic target recognition models. The soft segmented randomization framework applies a Gaussian mixture model to segment target and clutter regions softly, introducing randomized variations that align the synthetic data's statistical properties more closely with those of real-world data. Experimental results demonstrate that the proposed soft segmented randomization framework significantly enhances model performance on measured synthetic aperture radar data, making it a promising approach for robust automatic target recognition in scenarios with limited or no access to measured data.
\end{abstract}


\section{Introduction}
\label{sec:introduction}
Synthetic aperture radar (SAR) is remote sensing technology that, unlike electro-optical imaging, uses long wavelength bands to obtain high-resolution images regardless of time of day or weather conditions. This makes SAR widely used in various applications such as military and civilian surveillance, geological surveys, and more. However, the process of acquiring SAR data is very complex and costly, making it difficult to obtain experimental data. Specifically, it is nearly impossible to directly acquire SAR data of specific military targets, such as foreign tanks or aircraft.

\begin{figure*}[t!]
\centering
\includegraphics[width=0.85\linewidth]{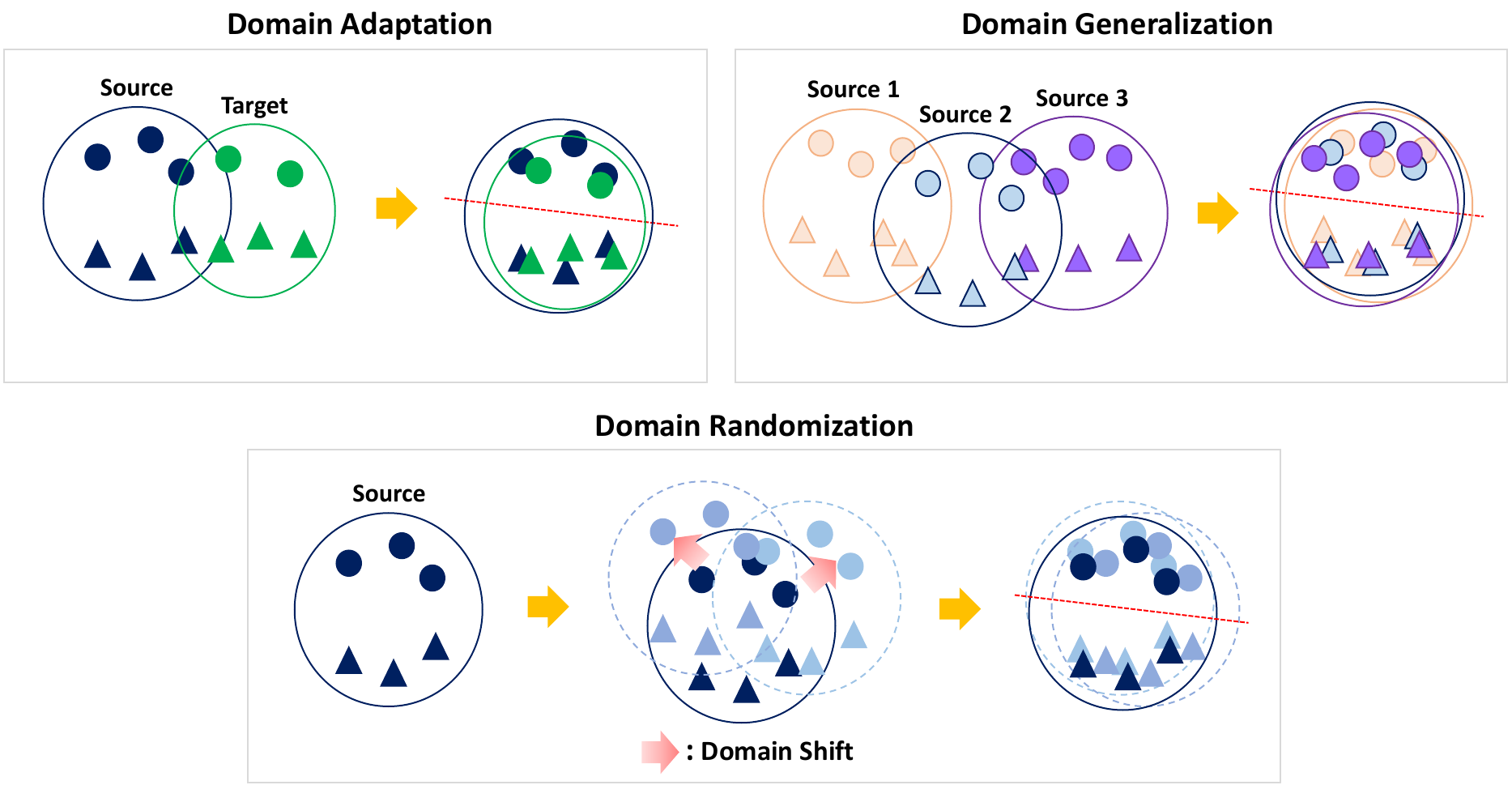}
\captionsetup{margin=1.3cm}
\caption{Domain adaptation, generalization and randomization}
\label{fig1}
\end{figure*}

To address these issues, a data acquisition method using CAD models and electromagnetic wave simulations has been proposed~\cite{ref1, ref2, ref3, ref4}. The simulation begins by creating realistic CAD models of the targets and estimating the reflective properties of each surface of the model and background. A ray-tracing based technique is typically used to approximate and predict the radar reflection signals at regular positions around a dome-shaped pattern surrounding the target~\cite{ref3, ref4, ref48, ref49}. The predicted reflection signals can be used to construct simulated phase records similar to those collected by sensors under normal operating conditions. This approach can significantly reduce the cost of data acquisition. Additionally, for certain military targets from foreign countries, where it is impossible to acquire measured data and no 3D drawings are available, it is possible to generate 3D drawings using indirect information (e.g., 2D drawings, multi-view photos, video clips) and obtain SAR data through electromagnetic simulation~\cite{ref27}.

However, differences between the synthetic and measured data arise due to inaccuracies in CAD models, errors in RCS calculations due to approximations, and differences from the real background clutter. These differences can degrade the performance of deep learning-based Automatic Target Recognition (ATR) networks trained solely on synthetic data~\cite{ref4}. This is because deep learning relies heavily on statistical methods based on the data, and tends to overlook out-of-distribution (OOD) scenarios commonly encountered in the real world, making it highly dependent on the assumption of identical distribution. As a result, deep learning models are robust only to the distribution of the training data and are vulnerable to domain discrepancy~\cite{ref21}. Nevertheless, deep learning, with its robust feature extraction and combination capabilities in convolutional neural networks (CNNs) based algorithms~\cite{ref5, ref6, ref7, ref8}, is superior to traditional ATR algorithms in terms of performance and generalizability~\cite{ref50}. If a deep learning network trained on synthetic data could be made robust against OOD scenarios, it could recognize targets in the real world without the need for measured SAR data, thereby reducing domain discrepancies between synthetic and measured data for critical tasks.

For optical deep learning, two main approaches for reducing domain discrepancies are domain adaptation (DA) and domain generalization (DG). The source domain refers to a domain in which data acquisition is easy, and large amounts of labeled data are available, whereas the target domain refers to a domain in which data acquisition is difficult, and only a small amount of labeled or unlabeled data is available~\cite{ref12, ref13, ref14, ref15, ref21}. In this context, synthetic data generated through simulations can be considered the source domain, whereas real-world measured data can be considered the target domain~\cite{ref22}. Recently, SAR-ATR researchers have leveraged methods developed in the optical domain to reduce domain discrepancy. DA is an approach that involves training with both the source domain data and a small amount of the target domain data, using methods that align the two domains at the feature level~\cite{ref22, ref23, ref24, ref25, ref26} or using generative models to transform the domains at the image level~\cite{ref18, ref19, ref20}. Although previous studies have achieved high classification accuracy in the target domain, they still rely on using target domain knowledge during training, which is a significant limitation. This remains a challenge for targets where data acquisition is costly or impossible.

In this study, we focus on DG. DG involves training on only one or multiple source domains without using a target domain, i.e., using 100\% synthetic data for training. This is a highly challenging task but offers the significant advantage of not requiring measured data for training. Although general DG approaches in optics involve using multiple source domains~\cite{ref21}, obtaining multiple source domains in SAR-ATR is highly challenging. Therefore, we use domain randomization (DR), a method within DG that operates with a single source domain. DR involves randomizing as many variables as possible, except for domain-agnostic features, to enhance the network's robustness to OOD scenarios not encountered during training. This augmentation technique allows the trained network to adapt well not only to synthetic data but also to various conditions and variability, such as clutter in real-world scenarios. Further, DR-generated data can be effectively used in cases where it is difficult to collect a large amount of data in real-world environments. Fig.~\ref{fig1} shows the difference between DA and general DG and DR. DR intentionally generates domain shift and aligns the shifted domains.

Given the above background, we aim to achieve high performance for a deep learning-based SAR-ATR network trained solely on synthetic data using DR against measured data. We believe that clutter, noise, and target signature elements at the image level of SAR data are the primary causes of domain discrepancies. Differences in clutter and noise are statistically reflected as differences in the mean and variance of image clutter, and target signature differences are visually prominent.

\begin{figure}[t]
\centering
\includegraphics[width=0.6\linewidth]{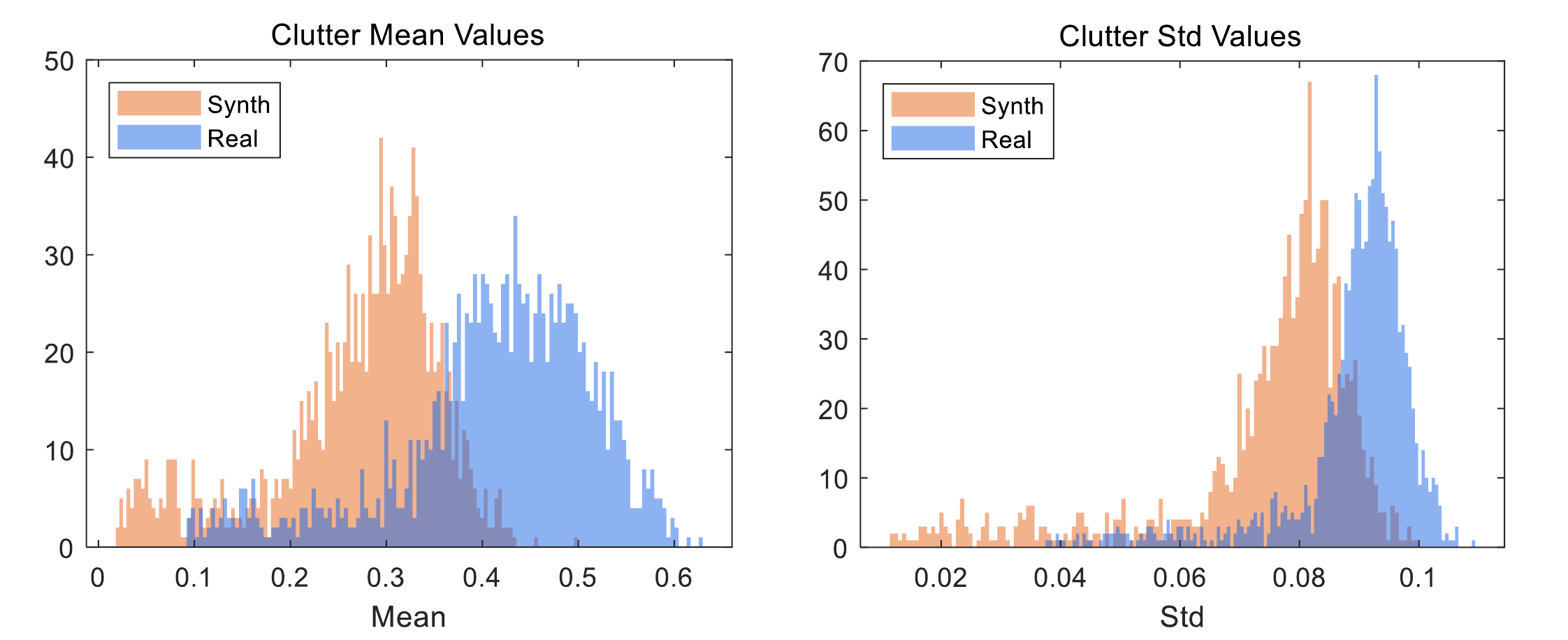}
\caption{Histograms of image-level clutter means and standard deviations}
\label{fig2}
\end{figure}

\textbf{Clutter Mean and Variance Differences:} In SAR-ATR, the sizes and shapes of targets, including shadows, are irregular and vary, making it difficult to isolate targets in images. Thus, clutter information is heavily included in images, and unrealistic synthetic data clutter leads to domain discrepancies. Statistically, as shown in Fig.~\ref{fig2}, domain discrepancy is reflected in differences in mean and variance~\cite{ref30}, which can reduce performance in deep learning that relies on pattern recognition~\cite{ref43}. Therefore, the network should be resilient to the statistical properties of the images, such as mean and variance. Differences in mean and variance can be interpreted as clutter reflectivity, and speckle noise intensity discrepancies, respectively.

\begin{figure}[t]
\centering
\includegraphics[width=0.5\linewidth]{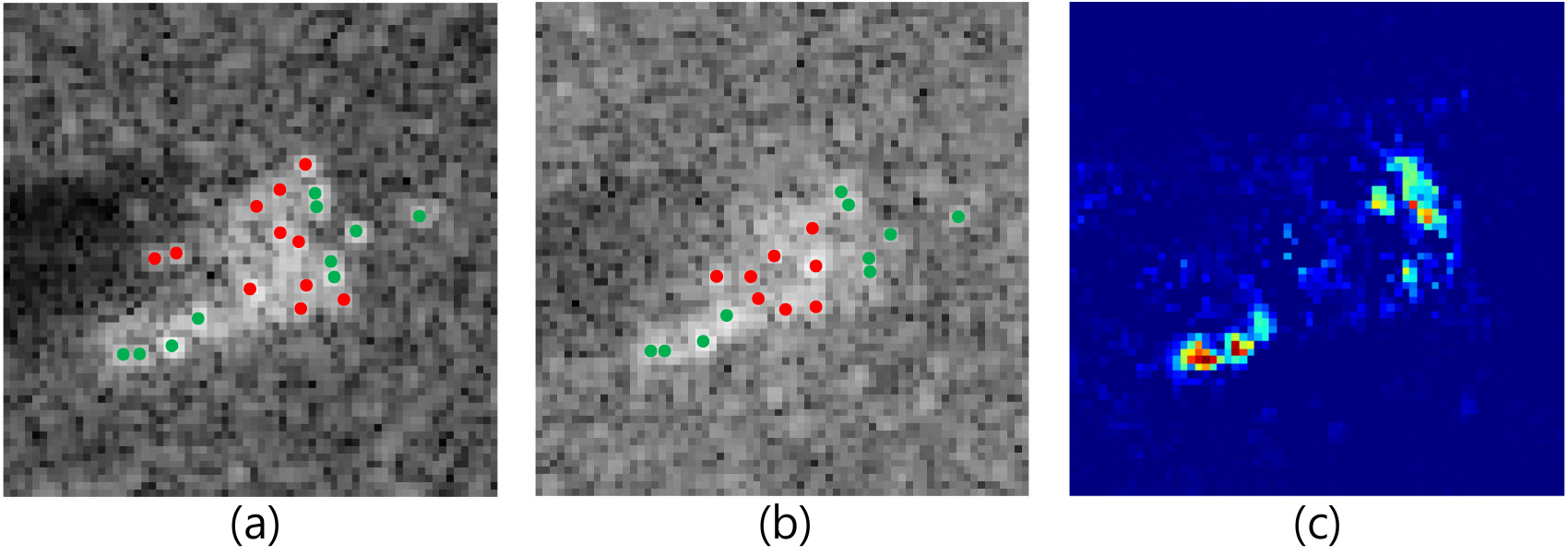}
\caption{Green points are similar location scattering topological points in the two domains, and red points are different location scattering topological points: (a) synthetic image, (b) measured image, (c) using the proposed method, the decision points for (b) are visualized using SHAP \cite{ref44} for a network trained with randomized images.}
\label{fig3}
\vspace{-1.2em}
\end{figure}

\textbf{Target Signature Differences:} Target signatures contain critical target information central to SAR-ATR. However, slight differences in the target signatures between domains, particularly scattering topological points, can lead to domain discrepancies. Therefore, using target signatures directly from synthetic data when training a network can cause domain discrepancies. However, if the CAD models used do not significantly differ from the real targets' characteristics, domain-agnostic features, such as certain scattering topological points with high intensity or large forms and specific aspects of target shapes, are common in both domains~\cite{ref22}. As shown in Fig.~\ref{fig3}, the network's decision points are primarily located around positions with high intensity or large forms among the green points that indicate similar locations of scattering topological points as well as in certain target shapes. It is crucial to maintain these features while modifying other target signatures.

To address the above issue, we propose a soft segmented randomization (SSR) framework. SSR applies a Gaussian mixture model (GMM) to a histogram to softly segment clutter and target regions based on their intensity distribution characteristics, applying different variations to each region. We mainly adjust mean and variance values to alter reflectivity and noise in the clutter region, while aiming to preserve as many domain-agnostic features as possible in the target region. By augmenting only synthetic data using SSR, we train a network and demonstrate through experiments that the proposed framework significantly reduces domain discrepancy and achieves high performance on measured data.

The major contributions of this study are summarized as follows.
\begin{enumerate}
    \item \textbf{Identification of Image-Level Domain Discrepancies}: We thoroughly analyze the primary causes of domain discrepancies in SAR-ATR at the image level, particularly focusing on differences in clutter mean and variance as well as target signature variations between synthetic and measured SAR data. These insights are critical for understanding how these discrepancies impact the performance of deep learning models.

    \item \textbf{Development of SSR}: To address the identified discrepancy, we propose SSR, which uses GMM for softly segmenting SAR images into the clutter and target regions. The SSR framework introduces targeted randomization to mitigate clutter reflectivity and speckle noise discrepancies while preserving essential domain-agnostic features in the target regions.

    \item \textbf{Enhanced DG for SAR-ATR}: The SSR framework significantly enhances the generalizability of deep learning models trained solely on synthetic data, enabling them to perform robustly on real-world SAR images without relying on measured data during training.

    \item \textbf{Practical Application to SAR-ATR Using Synthetic Data}: This study offers a practical solution for SAR-ATR systems that depend on synthetic data, especially in scenarios where acquiring real-world data is challenging or impossible. The SSR framework effectively bridges the gap between synthetic and measured data, enhancing the reliability and applicability of SAR-ATR in diverse operational environments.
\end{enumerate}

The remainder of this article is organized as follows.
Section \uppercase\expandafter{\romannumeral2} describes related work, datasets used in experiments, and the preprocessing steps. 
Section \uppercase\expandafter{\romannumeral3} details the proposed SSR framework for DR.
Section \uppercase\expandafter{\romannumeral4} presents the experimental results, demonstrating the reduction in domain discrepancy, along with an analysis.
Section \uppercase\expandafter{\romannumeral5} validates the effectiveness of the proposed method through ablation studies.
Finally, Section \uppercase\expandafter{\romannumeral6} concludes the paper.

\section{Background}
\subsection{Related Work Using Only Synthetic Data}
Most of the latest SAR-ATR algorithms require massive labeled data to achieve high accuracy. However, in several application scenarios, collecting and labeling massive of data for new tasks is unrealistic, particularly in SAR-ATR. This challenge can be greatly simplified if only synthetic data generated through simulations are used.

In SAR-ATR, various approaches have been proposed using only synthetic data to address the DG problem. For example, \cite{ref32} proposed a two-stage recognition algorithm that reclassifies data with significant domain differences based on reliability. This method involves multi-similarity fusion based on template matching, which is computationally expensive. In \cite{ref31}, only ensemble learning was employed, whereas \cite{ref30} combined ensemble learning with techniques such as Gaussian noise addition, dropout, and label smoothing, which have been employed in the optical domain to enhance model generalizability. Although they achieved high performance, the use of multiple networks during inference increases computational costs.

In another approach, \cite{ref17} performed DR by varying clutter reflectivity through contrast adjustment, whereas \cite{ref16} further diversified data by altering the number of ray bounces in simulations, adding noise, and introducing phase errors. Contrast adjustment may be effective in preserving high-intensity scattering topological points; however, because it is applied uniformly across an entire image, it fails to preserve relatively low-intensity but large-form scattering topological points. Moreover, some simulation-based methods show decreased performance, and even those with improvements exhibit only marginal gains, indicating that randomization through simulations does not adequately reflect variables encountered in real-world scenarios.

\subsection{Dataset}
\subsubsection{Synthetic and Measured Paried-Labeled Experiment (SAMPLE) Dataset}
To evaluate whether domain discrepancy has been reduced, we use the SAMPLE dataset released by the Air Force Research Laboratory (AFRL) in 2019~\cite{ref28}. SAMPLE was proposed to address the limitations of acquiring real-world data and has been widely used to study differences between synthetic and measured data \cite{ref17, ref18, ref20, ref22, ref23, ref24, ref25}.

\begin{figure*}[t]
\centering
\includegraphics[width=1\linewidth]{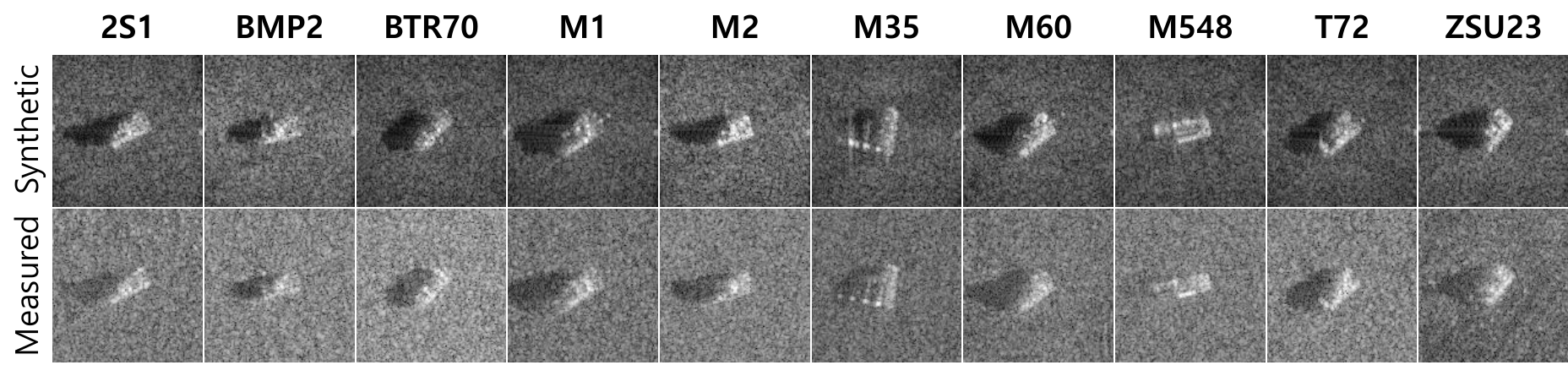}
\captionsetup{margin=0.15cm}
\caption{Synthetic and measured image pairs in SAMPLE dataset}
\label{fig4}
\end{figure*}

SAMPLE comprises approximately 1,300 synthetic and measured data pairs across 10 classes. It includes target classes overlapping with the moving and stationary target acquisition and recognition (MSTAR) dataset, and some measured data in SAMPLE are sourced from the MSTAR dataset. Each synthetic data instance in SAMPLE is generated using the same parameters as those used to simulate MSTAR targets with a range resolution of 0.3 m, image size of 128$\times$128 pixels, azimuth angle range of 10$^\circ$ to 80$^\circ$, and elevation angle range of 14$^\circ$ to 17$^\circ$. However, as shown in Fig. \ref{fig4}, the background of SAMPLE synthetic data is modeled as a homogeneous stochastic rough surface, which leads to noticeable differences in speckle noise intensity and clutter reflectivity compared with the measured data.

\subsubsection{MSTAR Dataset}

The MSTAR dataset was collected and released by AFRL and the Defense Advanced Research Projects Agency in the late 1990s~\cite{ref29}. MSTAR remains one of the most widely used datasets in SAR-ATR research, containing more than 8,000 target images and 100 clutter images. Similar to SAMPLE, the target data are provided across 10 classes (with partial overlap in classes), with a range resolution of 0.3 m and an image size of 128$\times$128 pixels. However, MSTAR offers a wider azimuth angle range and provides data at multiple elevation angles, including 15$^\circ$, 17$^\circ$, 30$^\circ$, and 45$^\circ$, making it more varied and extensive.

The target data from MSTAR are used to evaluate the effectiveness of the proposed method as a data augmentation technique. In addition, the clutter data, which also have a range resolution of 0.3 m but are provided at a larger image size of 1,700$\times$1,400 pixels and an elevation angle of 15$^\circ$$-$16$^\circ$, are used in conjunction with SAMPLE and MSTAR target data to assess how well the proposed method mitigates the impact of clutter on network performance.

\subsection{Dynamic Range Reduction}
The single-look SAR data used here includes both amplitude and phase information, providing the reflectivity information of each resolution cell. Reflectivity is strongly influenced by the backscatter RCS of scenes, which is affected by the point of view and nature of the backscatterer. Such RCS can have significant variations depending on the measured data, resulting in a very large dynamic range. Even if the measured and synthetic data are acquired under the same target and operational conditions, differences in dynamic range may arise. These differences can significantly impact the performance of deep learning-based SAR-ATR, which relies on statistical data. Therefore, preprocessing techniques that include dynamic range handling must be aligned.

In this study, to eliminate the dynamic range differences between measured and synthetic data, we use logarithm mapping, the most commonly used technique for this purpose \cite{ref34, ref35}. This technique converts speckle noise, which has multiplicative characteristics, into additive noise as follows:

\begin{equation}
    I = \frac{\log_{10}(1 + cA)}{\log_{10}(1 + c)},
\end{equation}
where \( A \) represents the min-max normalized amplitude value, \( c \) is a parameter that controls brightness, and $I$ denotes the intensity image with values in the range [0, 1]. In this study, we set \( c \) to 1,000.

\section{Proposed Method}
Before explaining the proposed method, we first describe the formulation of DR. Denote $\mathcal{X}$ and $\mathcal{Y}$ as the input (feature) and label spaces, respectively. A domain defined on $\mathcal{X}\times\mathcal{Y}$ can be represented by a joint probability distribution $\mathbb{P}(X,Y)$. The set $X$ comprises individual data instances, denoted as $X=\{I^1, I^2, \ldots, I^N\}$, whereas $Y=\{y^1, y^2, \ldots, y^N\}$ represents the corresponding labels. 

In the context of DR, we assume access to only a sufficiently large labeled single source domain $\mathcal{D}^S = \{X^S, Y^S\}$, with each domain having a joint distribution $\mathbb{P}^S(X,Y)$. The goal of DR is to ensure that a prediction model $f:X \rightarrow Y$, trained solely on $\mathcal{D}^S$, produces low prediction errors on an unseen target domain $\mathcal{D}^T = \{X^T\}$. Here, $\mathbb{P}^S(X,Y) \neq \mathbb{P}^T(X,Y)$.

We assume that \( X \) comprises domain-agnostic features \( Z_{da} \), which contain common information across multiple domains, and domain-specific features \( Z_{ds} \), which are unique to a particular domain. Thus, \( \mathbb{P}^S(X,Y) \) can be redefined as \( \mathbb{P}^S(Z_{da}, Z_{ds}, Y) \). The key to DR is to remove \( Z_{ds} \), thereby preventing the model from learning domain-specific dependencies, and ensuring that it generalizes across diverse domains. This can be simplified by randomizing and marginalizing \( Z_{ds} \) as follows:

\begin{equation}
    \mathbb{P}^S(Z_{da}, Y) = \int \mathbb{P}^S(Z_{da}, Z_{ds}, Y) dZ_{ds}
\end{equation}

By eliminating \( Z_{ds} \), which causes domain discrepancies, the network relies solely on \( Z_{da} \), enabling consistent performance across various domains. 
We believe that \( Z_{ds} \) is closely related to the statistical differences in the clutter and target signatures, as previously argued. Therefore, to explicitly randomize \( Z_{ds} \) at the image level, we propose SSR.

\begin{figure}[t]
\centering
\includegraphics[width=0.95\linewidth]{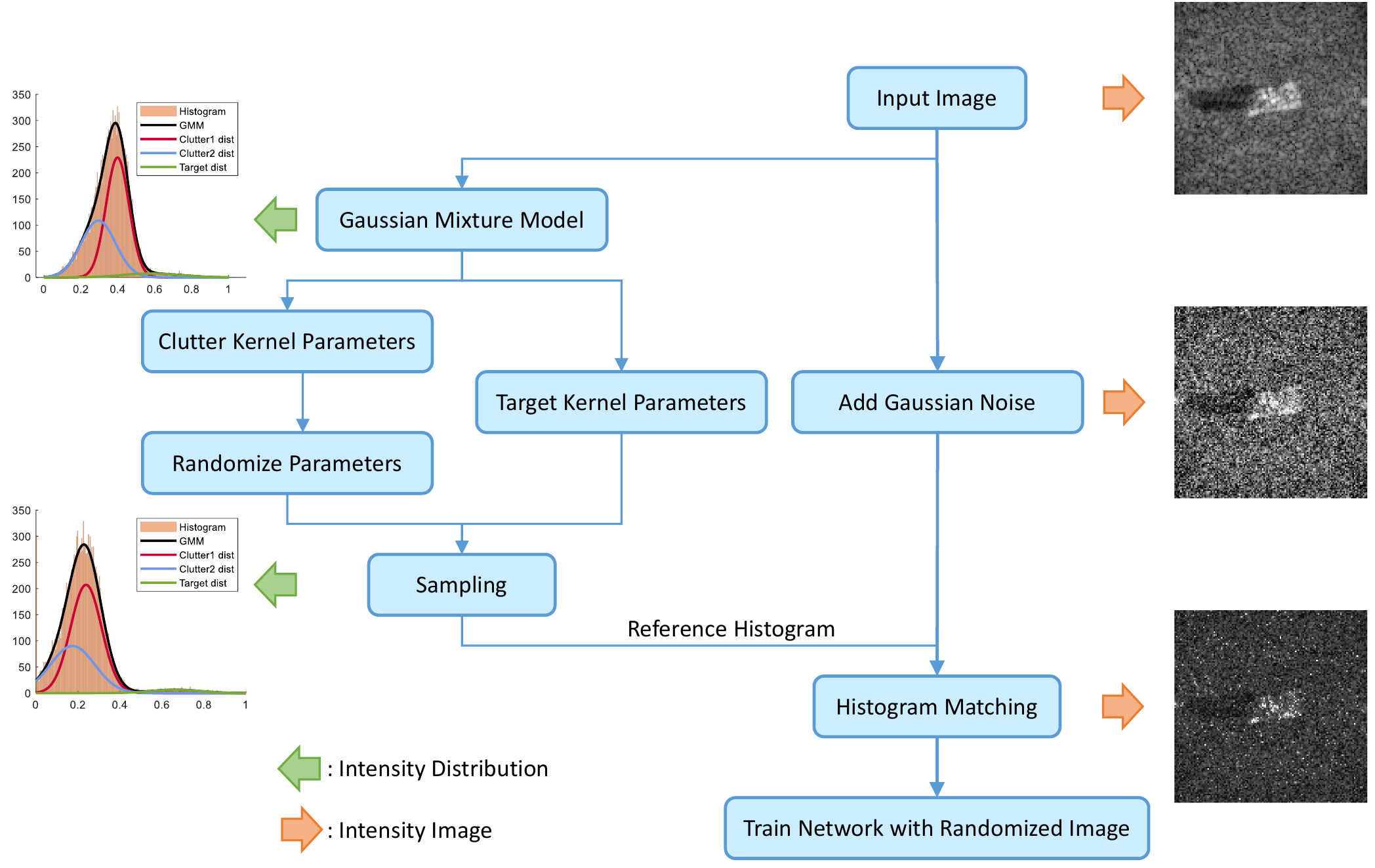}
\captionsetup{margin=0.7cm}
\caption{Flow diagram of the SSR framework}
\label{fig13}
\end{figure}

As shonw in Fig.~\ref{fig13}, SSR is a DR augmentation framework comprising three main steps designed to address DG issues. First, the intensity distribution of each synthetic SAR image is modeled using a GMM to obtain the kernel parameters. Next, noise is added to introduce variations in speckle noise and target signatures. Finally, variations in intensity are introduced by adjusting the mean and variance of the kernel parameters. This section elaborates on the proposed method in detail.

\subsection{Gaussian Mixture Modeling}
In X-band SAR images, the intensity in target regions typically has higher values than that in clutter regions because of the higher RCS of targets \cite{ref45}. Using this characteristic, we apply a GMM to the image histogram using three Gaussian kernels to softly segment the target and clutter regions effectively.

GMM is a machine learning algorithm that models complex probability distributions using multiple Gaussian kernels, each defined by parameters such as mean, variance, and mixing coefficient \cite{ref46}. The mixing coefficient represents the proportion of data modeled by each Gaussian distribution and determines the contribution of each Gaussian to the overall model.

\begin{figure*}[t]
\centering
\includegraphics[width=0.6\linewidth]{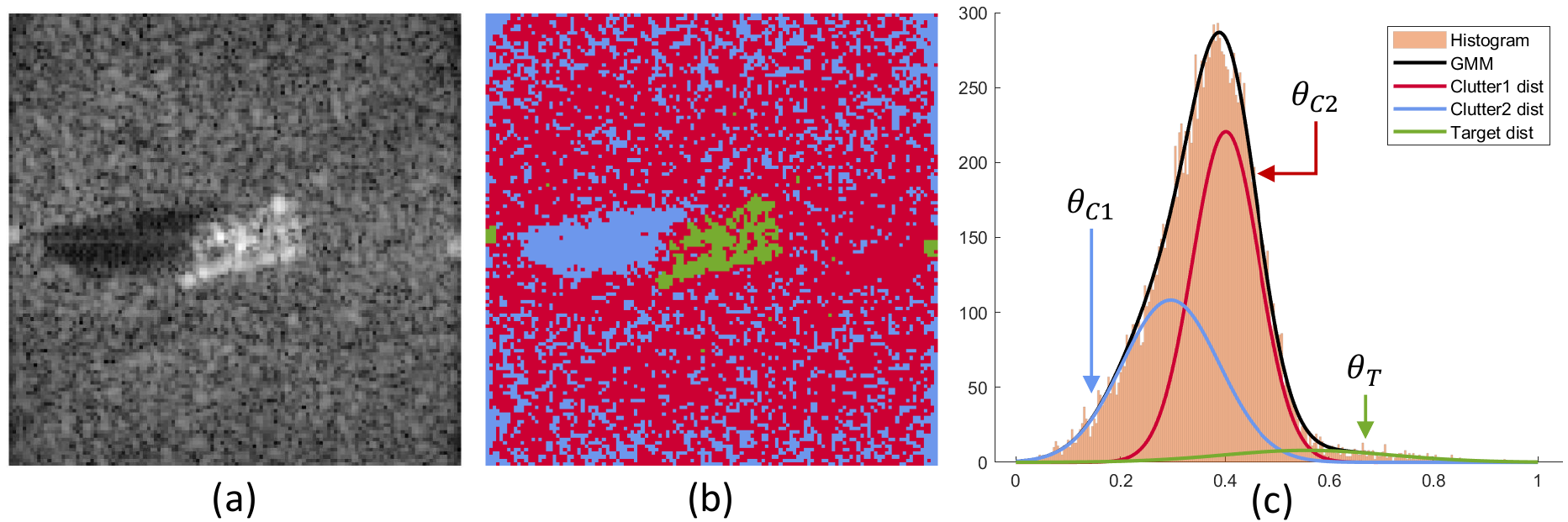}
\caption{(a) SAMPLE synthetic image, (b) result of segmenting (a) using a GMM based on the intersections of the Gaussian kernels in (c). (c) Gaussian kernels obtained by histogram and GMM for (a). The blue and red Gaussian kernels $\theta_{C1, C2}$ in (c) represent the clutter and shadow regions. The green Gaussian kernel $\theta_{T}$ in (c) represents the target regions and the strong scatter points commonly appearing on both sides in SAMPLE synthetic images.}
\label{fig5}
\end{figure*}

\begin{figure}[t]
\centering
\includegraphics[width=0.4\linewidth]{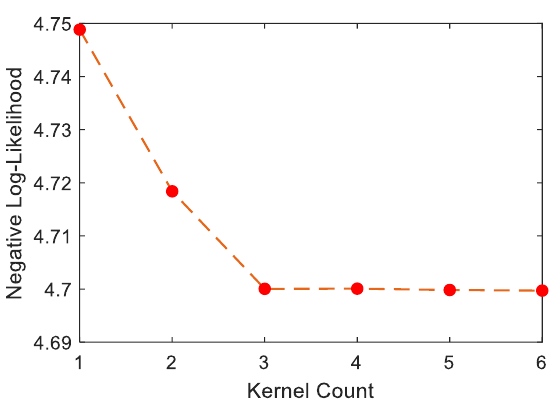}
\caption{Negative log likelihood after convergence according to Gaussian kernel count. The kernel count exceeding three does not lead to further convergence.}
\label{fig6}
\end{figure}

Gaussian mixture modeling offers an effective approach to modeling the intensity distribution of targets and clutter in complex SAR images. SAR simulators typically use simple clutter models, such as homogeneous stochastic rough surfaces or grass, because simulating complex surfaces is challenging \cite{ref3, ref4, ref35}. In addition, the speckle noise is simplified using a multiplicative mechanism, which makes the scattering distribution follow a Gaussian distribution \cite{ref35}. Because of these factors, the intensity distribution in the clutter region tends to follow a Rayleigh distribution. We transform the Rayleigh distribution into a log-Rayleigh distribution using logarithm mapping \cite{ref34}, which can be easily modeled using two Gaussian kernels \cite{ref36}. The remaining Gaussian kernel is used to model the intensity distribution of the target regions. As a result, the target and clutter regions are softly segmented and modeled using three Gaussian kernels (see Fig.~\ref{fig5} and~\ref{fig6}).

To obtain the distribution modeled by these the Gaussian functions, the expectation-maximization (EM) algorithm is employed. Suppose the intensity histogram of a SAR image is divided into $M$ bins, where the center value of each bin is denoted as $B = \{b_1, b_2, \ldots, b_m, \ldots, b_M\}$. The mean, standard deviation, and mixing coefficient of the $k$th Gaussian kernel are represented as $\mu_k$, $\sigma_k$, and $\pi_k$, respectively.

\begin{equation}
    \mathcal{N}(b_m|\mu_k, \sigma_k^2) = \frac{1}{\sqrt{2\pi\sigma_k^2}} \exp\left(-\frac{(b_m - \mu_k)^2}{2\sigma_k^2}\right)
\end{equation}

\begin{equation}
    p(b_m) = \sum_{k=1}^{3} \pi_k \cdot \mathcal{N}(b_m|\mu_k, \sigma_k^2)
\end{equation}

The EM algorithm optimizes these parameters and can be summarized as follows:

\begin{enumerate}
    \item Initialize $\theta = \{\pi_1, \pi_2, \pi_3, \mu_1, \mu_2, \mu_3, \sigma_1, \sigma_2, \sigma_3\}$.
    \item Evaluate posterior probability \( \gamma \) using the current parameter values:
    \begin{equation}
    \gamma_k(b_m) = \frac{\pi_k \mathcal{N}(b_m | \mu_k, \sigma_k^2)}{\sum_{k=1}^{3} \pi_k \mathcal{N}(b_m | \mu_k, \sigma_k^2)}
    \end{equation}
    \item Re-estimate the parameters \( \theta \) using the current responsibilities:
    \begin{equation}
        \begin{aligned}
    \mu_k^{\text{new}} = \frac{\sum_{m=1}^{M} \gamma_k(b_m) b_m}{\sum_{m=1}^{M} \gamma_k(b_m)}\\
    (\sigma_k^{\text{new}})^2 = \frac{\sum_{m=1}^{M} \gamma_k(b_m) (b_m - \mu_k^{\text{new}})^2}{\sum_{m=1}^{M} \gamma_k(b_m)}\\
    \pi_k^{\text{new}} = \frac{1}{M} \sum_{m=1}^{M} \gamma_k(b_m)
        \end{aligned}
    \end{equation}
    \item Calculate the negative log-likelihood and determine whether either the negative log-likelihood or parameter values have converged. If the convergence criterion is not met, return to step 2.
    \begin{equation}
    L(B;\theta) = -\sum_{m=1}^{M} \log \left( \sum_{k=1}^{3} \pi_k \mathcal{N}(b_m | \mu_k, \sigma_k^2) \right)
    \end{equation}
\end{enumerate}

As a result of the optimization, three Gaussian kernels are obtained with the parameters $\theta = \{\theta_{C1}, \theta_{C2}, \theta_{T}\}$, sorted in ascending order of mean values, where $\theta_{C1} = \{\mu_{C1}, \sigma_{C1}, \pi_{C1}\}$, $\theta_{C2} = \{\mu_{C2}, \sigma_{C2}, \pi_{C2}\}$, and $\theta_{T} = \{\mu_{T}, \sigma_{T}, \pi_{T}\}$.

\subsection{Noise}
Speckle noise is a complex type of noise that arises from phase interference between sub-resolution objects due to the coherent nature of radar systems. Although it is difficult to analyze, simulators simplify speckle noise using a multiplicative mechanism; because we use logarithm mapping, speckle noise can be represented as an additive noise model as follows~\cite{ref35}:
\begin{equation}
    I = R + n_s,
\end{equation}
where $R$ represents the RCS of a scene, and $n_s$ is the speckle noise following a log-Rayleigh distribution. The intensity of speckle noise in the clutter region of the synthetic data is lower than that of the measured data, which manifests as a difference in variance. If Gaussian noise is not added, the network might overfit by also learning the clutter. To introduce variations in the distribution of clutter speckle noise, Inkawhich et al.~\cite{ref30} added Gaussian noise, which helped their network to focus more on target signatures, thereby improving the accuracy of measured data.

Because Gaussian noise is applied across an entire image, it also affects target signatures. Among target signatures, domain-agnostic features such as some scattering topological points with high intensity or large forms are less distorted than other regions with domain-specific features of lower intensity and smaller forms. This helps the network focus on less distorted regions and learn domain-agnostic features. By adding Gaussian noise, we enhance the network's focus on target signatures while introducing appropriate variations in target signatures to improve generalizability:
\begin{equation}
    I_{\text{noise}} = R + n_s + n'_s,
\end{equation}
where $I_{\text{noise}}$ denotes the SAR intensity image with added noise, and $n'_s$ denotes the Gaussian noise.

\begin{figure*}[t!]
\centering
\includegraphics[width=0.7\linewidth]{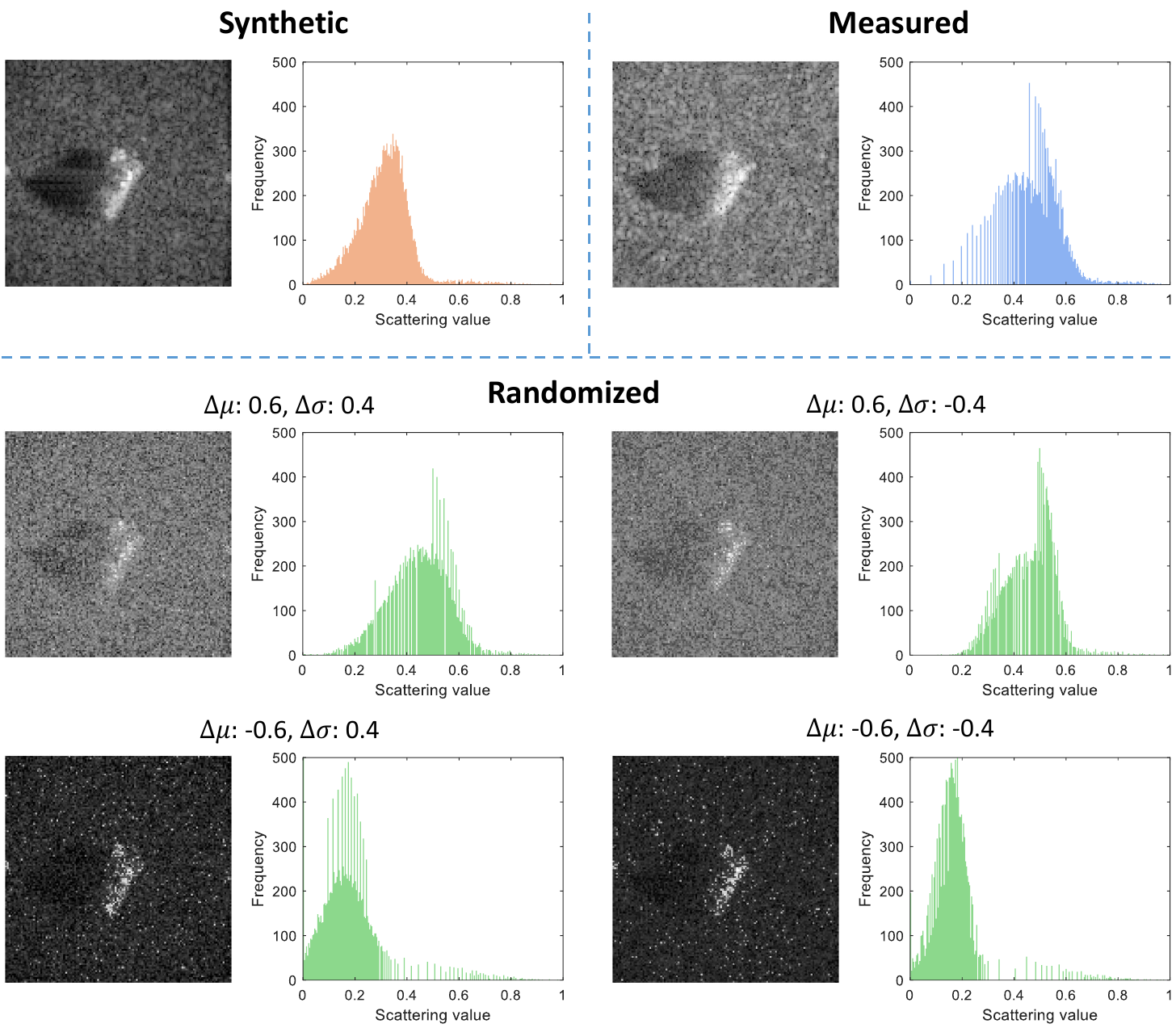}
\caption{The upper left and upper right are the synthetic and measured data and their histograms, respectively. Below are the randomized(augmented) data using SSR and their histograms.}
\label{fig7}
\end{figure*}

\subsection{GMM-based Adjustment}
The data with added Gaussian noise undergoes changes in both speckle noise and target signatures. However, the statistical properties still differ from those of the measured data, which can affect the performance of deep learning networks that rely on pattern recognition. This indicates that the network cannot be considered fully generalized. We address this issue using a GMM-based approach.

Among the three Gaussian kernels derived, $\theta_{C1}$ and $\theta_{C2}$ primarily contain clutter intensity information, whereas $\theta_{T}$ mostly includes target intensity information. As previously identified, differences in the mean and variance of the clutter region, as well as differences in target signatures, are key contributors to domain discrepancies. Therefore, we adjust the mean and standard deviation parameters of $\theta_{C1}$ and $\theta_{C2}$, which contain significant clutter intensity information, to change the histogram distribution of clutter intensity and reduce the statistical differences. Although this adjustment also affects shadow regions, it does not distort the shape of these regions. Because segmentation is applied softly, parts of the target signatures with lower or similar intensities than the clutter are also adjusted during this process. This helps the network focus more precisely on essential and refined target signatures.

The Gaussian kernel $\theta_{T}$, which contains target intensity information, is not changed. This kernel primarily holds crucial information about scattering topological points, and its careful handling is necessary. The experimental results indicated that minor changes to this kernel did not yield significant differences, whereas substantial changes led to the loss of scattering topological point information, ultimately reducing performance. Thus, we chose not to modify $\theta_{T}$.

\begin{equation}
    \begin{aligned}
        \mu_{C1,C2}^{\prime} = \mu_{C1,C2} \times (1 + \Delta\mu), \quad \Delta\mu \sim \mathcal{U}(-\alpha, \alpha)\\
        \sigma_{C1,C2}^{\prime} = \sigma_{C1,C2} \times (1 + \Delta\sigma), \quad \Delta\sigma \sim \mathcal{U}(-\beta, \beta)\\
        p_{\text{adj}}(b_m) = \sum_{k=1}^{2} \pi_k \cdot \mathcal{N}(b_m \mid \mu_k', \sigma_k'^2) + \pi_3 \cdot \mathcal{N}(b_m \mid \mu_3, \sigma_3^2)
    \end{aligned}
\end{equation}
where $\alpha$, and $\beta$ denote the change rates of the mean and standard deviation of the Gaussian kernels. After these adjustments, we sample new data based on the adjusted distribution, maintaining the same size as the synthetic image. The final image is then obtained through histogram matching with the noise-added image, followed by clipping the values to the range [0, 1].

\begin{equation}
    \begin{aligned}
    I_{\text{sampled}} &\sim p_{\text{adj}}(b_m)\\
    I_{\text{rand}} = H^{-1}_{\text{noise}} &\left(H_{\text{sampled}}(I_{\text{sampled}})\right)
    \end{aligned}
\end{equation}
where $I_{sampled}$ denotes an image sampled from the adjusted distribution, $I_{rand}$ denotes the randomized final image, $H_{noise}$ denotes the cumulative distribution function (CDF) of $I_{noise}$, and $H_{sampled}$ denotes the CDF of $I_{sampled}$. The randomized final image adjusts speckle noise, clutter reflectivity, and target signatures. By repeatedly randomizing $\alpha$ and $\beta$ to augment new data, we can prevent the network from learning the causes of domain discrepancy identified at the image level. In each iteration, new images are augmented based on randomly adjusted histogram distributions (Fig.~\ref{fig7}), ensuring that the network avoids overfitting to specific clutter patterns or target signatures. This randomization process suppresses domain-specific features and facilitates the network to focus on domain-agnostic features, thereby improving its generalizability.

The intensity distribution of a randomized image is not guaranteed to follow a log-Rayleigh distribution. In the real world, clutter and speckle noise mechanisms are complex, leading the scattering distribution to deviate from a Gaussian distribution; this means that the intensity distribution does not precisely follow a log-Rayleigh distribution \cite{ref35, ref47}. Therefore, we do not impose strict constraints on the intensity distribution and instead randomly adjust the mean and variance to generate a distribution similar to that of log-Rayleigh. Consequently, this process ensures that the network does not learn fixed patterns but instead reflects the variability that may occur across different domains, leading to a robust model capable of handling diverse changes that can arise in real-world operational environments.

\begin{algorithm}
  \caption{Main algorithm of the proposed soft segmented randomization (SSR)}
  \begin{algorithmic}[1]
    \Inputs{training samples $X=\{I^1, I^2, \ldots, I^N\}$\\ corresponding label $Y=\{y^1, y^2, \ldots, y^N\}$ \\ noise standard deviation (std) $\sigma_s$ \\ rate of change in the mean of kernel $\alpha$ \\ rate of change in the std of kernel $\beta$ \\ deep learning ATR network $f$}
    \Outputs{\strut$w_i^0 \gets 0$, $i=1,\ldots,n$ \\ $S_0 \gets S$}
    \StepOne \Statex Gaussian Mixture Modeling
    \For{$I^n \in X$}
        \State Initialize $\theta^n=\{\mu^n_k, \sigma^n_k, \pi^n_k\}$ for $k=1,2,3$
        \State Compute histogram $B^n=\{b_1^n, b_2^n, \ldots, b_M^n\}$ from $I^n$
        \While{not converged}
            \State $\theta^n \leftarrow$ Update parameters using EM algorithm
        \EndWhile
    \EndFor
    \State \Return $\Theta = \{\theta^1, \theta^2, \ldots, \theta^N\}$
    \StepTwo \Statex Randomization
    \For{$I^n \in X, y^n \in Y, \theta^n \in \Theta$}
        \State $I^n_{noise} \leftarrow I^n + n'_s$ where $n'_s$ is gauss noise with std of $\sigma_s$
        \State $\Delta\mu\leftarrow$ randomly selected from $\mathcal{U}(-\alpha, \alpha)$
        \State $\Delta\sigma\leftarrow$ randomly selected from $\mathcal{U}(-\beta, \beta)$
        \State $\mu_{C1,C2}^n \leftarrow \mu_{C1,C2}^n \times (1 + \Delta\mu)$
        \State $\sigma_{C1,C2}^n \leftarrow \sigma_{C1,C2}^n \times (1 + \Delta\sigma)$ where $C1, C2$ are two low mean kernels
        \State $I_{sampled}^n \leftarrow$ Samples from the three kernel distribution
        \State $I_{rand}^n \leftarrow$ Map the intensity values of $I_{noise}$ to $I_{sampled}$
        \State Train the ATR network $f$ using $(I^n_{rand}, y^n)$
    \EndFor
    \State \Return trained network $f$
  \end{algorithmic}
\end{algorithm}

\section{Experiments and Results}
\subsection{Experimental Setup}
To evaluate the effectiveness of the proposed method in reducing domain discrepancy and enhancing model generalizability of networks, we consider eight CNN architectures: two CNNs specifically designed for SAR-ATR \cite{ref6, ref7} and well-known CNNs originally developed for optical imagery, i.e., VGG-11, VGG-16 \cite{ref37}, ResNet18, ResNet50 \cite{ref38}, MobileNet v2 \cite{ref39}, ShuffleNet v2 \cite{ref40}, AConvNet \cite{ref6}, and AM-CNN \cite{ref7}. These networks vary in size, with the number of parameters ranging from 0.3 million in AConvNet to 11.2 million in ResNet18 and 134 million in VGG-16.

AConvNet and AM-CNN are designed with input sizes of 88$\times$88 and 100$\times$100, respectively; thus, we crop images carefully to ensure that the targets and their shadows are not truncated. The other networks use an input size of 128$\times$128. We believe that preserving the shadow regions is crucial because they provide additional information about the targets \cite{ref5}. 

Given that the proposed method significantly adjusts the intensity distribution in the clutter region, one might question whether reducing the clutter region would diminish its effectiveness. However, experiments using AConvNet and AM-CNN demonstrate that the proposed method remains effective even when the clutter is reduced.

All networks are trained using the Adam optimizer for 100 epochs with 10 training and testing runs to calculate the mean accuracy and standard deviation. Depending on the network size, a batch size of 16 or 32 is used, and the learning rate is adjusted for each network. During training, the proposed randomized augmentation technique is applied with a 50\% probability. We set $\alpha$ to 0.6 and $\beta$ to 0.4 and applied Gaussian noise with a standard deviation of 0.3, which was selected after testing values of [0.1, 0.2, 0.3, 0.4, 0.5] to identify the optimal value.

\begin{table*}[]
\centering
\caption{Training and testing datasets under domain discrepancy reduction experimental scenarios}
\label{table1}
\resizebox{0.8\textwidth}{!}{%
\begin{tabular}{cccccc}
\noalign{\global\arrayrulewidth=1.0pt}
\hline
\noalign{\global\arrayrulewidth=0.4pt}
\multirow{2}{*}{\textbf{Class}} & \multirow{2}{*}{\textbf{Serial No.}} & \multicolumn{2}{c}{\textbf{First Scenario}} & \multicolumn{2}{c}{\textbf{Second Scenario}}     \\ \cline{3-6} 
                                &                                      & Train (14$^\circ$$-$16$^\circ$)        & Test (17$^\circ$)       & Train (14$^\circ$$-$17$^\circ$) & Test (14$^\circ$$-$17$^\circ$) + Clutter \\ \hline
2S1                             & B01                                  & 116                       & 58              & 174                & 174                         \\
BMP2                            & 9563                                 & 55                        & 52              & 107                & 107                         \\
BTR70                           & C71                                  & 43                        & 49              & 92                 & 92                          \\
M1                              & 0AP00N                               & 78                        & 51              & 129                & 129                         \\
M2                              & MV02GX                               & 75                        & 53              & 128                & 128                         \\
M35                             & T839                                 & 76                        & 53              & 129                & 129                         \\
M548                            & C245HAB                              & 116                       & 60              & 176                & 176                         \\
M60                             & 3336                                 & 75                        & 53              & 128                & 128                         \\
T72                             & 812                                  & 56                        & 52              & 108                & 108                         \\
ZSU23-4                         & D08                                  & 116                       & 58              & 174                & 174                         \\ \noalign{\global\arrayrulewidth=1.0pt} \hline
\end{tabular}
}
\end{table*}

\begin{figure}[hbt!]
\centering
\includegraphics[width=0.55\linewidth]{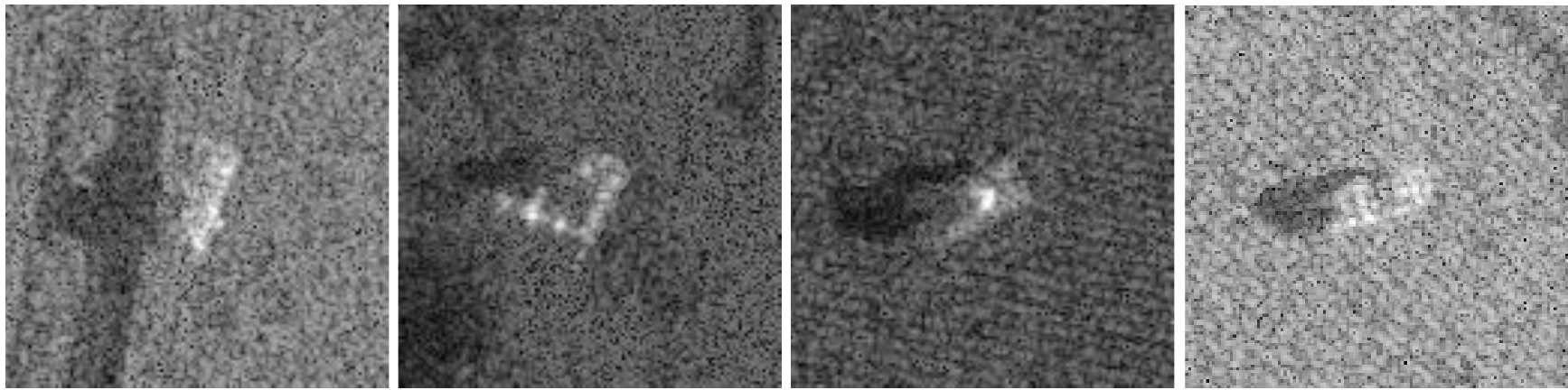}
\caption{SAMPLE measured data merged with MSTAR clutter}
\label{fig8}
\end{figure}

\subsection{Domain Discrepancy Reduction}
To evaluate the performance of the proposed randomized augmentation method, we conducted various comparative experiments. First, following \cite{ref30}, we applied strong Gaussian noise with a standard deviation of 0.4 and conducted comparative experiments to assess whether the proposed method outperforms simply adding strong noise. Moreover, \cite{ref16, ref17} reported performance improvements through contrast adjustment techniques, but the specific methods used were not detailed. Therefore, we combined the representative contrast adjustment method, gamma transformation, with Gaussian noise for comparison (after adding Gaussian noise, we applied gamma transformation). We also analyzed the effect of each component using only a GMM without Gaussian noise in the proposed method. This analysis serves as crucial evidence to verify whether the proposed method offers performance improvements beyond simple noise addition or contrast adjustment.

\subsubsection{First Scenario} In this scenario, only SAMPLE synthetic data with depression angles of 14$^\circ$, 15$^\circ$, and 16$^\circ$ were used for training. SAMPLE measured data with a 17$^\circ$ depression angle were used for testing (Table~\ref{table1}) to evaluate how effectively domain discrepancy can be overcome under standard operation conditions. Although simulations are much more cost-effective than acquiring measured data, all operational conditions cannot be considered perfectly. This scenario aims to verify whether a model trained on synthetic data can robustly operate under slight depression angle differences that may occur in real-world conditions.

The ATR results for each network are summarized in Table~\ref{table2}; the highest performance is highlighted in bold. \textbf{w/o Aug}, which is the case where only synthetic data are used, showed the lowest performance, indicating that models trained solely on synthetic data do not generalize well to measured data. \textbf{Only Noise}, which adds strong Gaussian noise, showed improved performance for all networks. \textbf{Gamma}, which combines contrast adjustment with noise, achieved higher and more stable performance than \textbf{Only Noise}, indicating that contrast adjustment can have a positive impact on DG. \textbf{SSR w/o Noise} which uses the proposed method without noise outperformed \textbf{Only Noise}. This demonstrates that adjusting the mean and standard deviation of the clutter region using GMM—i.e., adjusting the clutter reflectivity and speckle noise intensity—is more effective than merely adding noise. Finally, the \textbf{SSR} results show that the proposed method combined with noise achieved the highest and most stable performance. This confirms that the proposed framework effectively reduces domain discrepancy and significantly improves the model generalizability.

\subsubsection{Second Scenario} In this scenario, the entire SAMPLE synthetic dataset was used for training and the entire SAMPLE measured dataset merged with MSTAR clutter was used for testing (see fig.~\ref{fig8} and Table~\ref{table1}). Because the depression angles of the synthetic and measured datasets match, this scenario evaluates whether the data considered during the simulation phase can still perform effectively in real operational environments with diverse clutter. This allows us to assess network performance across various environments, beyond domain discrepancy issues.

\begin{table*}[]
\centering
\caption{The results of the first scenario experiments ATR accuracy [\%]}
\label{table2}
\resizebox{0.8\textwidth}{!}{%
\begin{tabular}{lccccc}
\noalign{\global\arrayrulewidth=1.0pt}
\hline
\noalign{\global\arrayrulewidth=0.4pt}
\multicolumn{1}{c}{} & \raisebox{0pt}[1.1em][0.5em]{\textbf{w/o Aug}}    & \textbf{Only Noise \cite{ref30}} & \textbf{Gamma \cite{ref16, ref17}} & \textbf{SSR w/o Noise} & \textbf{SSR (Proposed)} \\ \hline
VGG-11        & 39.67 $\pm$  2.29                  & 63.81 $\pm$  3.74                     & 84.42 $\pm$  1.91                         & 79.37 $\pm$  3.26                        & \textbf{87.51 $\pm$  0.89}              \\
VGG-16        & 39.78 $\pm$  2.84                  & 66.47 $\pm$  5.73                     & 83.62 $\pm$  2.22                         & 72.82 $\pm$  5.12                        & \textbf{87.64 $\pm$  1.52}              \\
ResNet18      & 52.45 $\pm$  6.21                  & 61.84 $\pm$  4.15                     & 90.33 $\pm$  4.50                         & 90.22 $\pm$  2.19                        & \textbf{94.69 $\pm$  1.19}              \\
ResNet50      & 36.23 $\pm$  7.73                  & 55.88 $\pm$  8.40                     & 85.34 $\pm$  4.63                         & 83.36 $\pm$  3.41                        & \textbf{92.95 $\pm$  2.22}              \\
MobileNet v2  & 43.27 $\pm$  7.05                  & 68.57 $\pm$  9.73                     & 87.77 $\pm$  2.40                         & 83.71 $\pm$  3.87                        & \textbf{91.97 $\pm$  1.98}              \\
ShuffleNet v2 & 36.68 $\pm$  4.87                  & 65.05 $\pm$  3.06                     & 88.72 $\pm$  2.31                         & 80.35 $\pm$  3.87                        & \textbf{91.95 $\pm$  1.30}              \\
AConvNet      & 55.70 $\pm$  12.41                  & 85.75 $\pm$  5.33                     & 88.01 $\pm$  2.86                         & 79.63 $\pm$  7.23                        & \textbf{91.15 $\pm$  2.59}              \\
AM-CNN        & 79.13 $\pm$  5.42                  & 79.92 $\pm$  6.34                     & 91.99 $\pm$  2.40                         & 90.06 $\pm$  2.49                        & \textbf{94.38 $\pm$  1.20}              \\ \noalign{\global\arrayrulewidth=1.0pt} \hline
\end{tabular}%
}
\end{table*}

\begin{table*}[]
\centering
\caption{The results of the second scenario experiments ATR accuracy [\%]}
\label{table3}
\resizebox{0.8\textwidth}{!}{%
\begin{tabular}{lccccc}
\noalign{\global\arrayrulewidth=1.0pt}
\hline
\noalign{\global\arrayrulewidth=0.4pt}
\multicolumn{1}{c}{} & \raisebox{0pt}[1.1em][0.5em]{\textbf{w/o Aug}}    & \textbf{Only Noise \cite{ref30}} & \textbf{Gamma \cite{ref16, ref17}}  & \textbf{SSR w/o Noise}  & \textbf{SSR (Proposed)} \\ \hline
VGG-11        & 58.95 $\pm$  3.58                  & 74.49 $\pm$  1.50                     & 84.98 $\pm$  0.38                         & 75.11 $\pm$  1.80                     & \textbf{87.60 $\pm$  1.20}              \\
VGG-16        & 53.52 $\pm$  6.84                  & 75.52 $\pm$  3.52                     & 85.55 $\pm$  1.42                         & 71.07 $\pm$  4.13                     & \textbf{88.05 $\pm$  1.60}              \\
ResNet18      & 79.68 $\pm$  4.08                  & 75.99 $\pm$  5.98                     & 91.67 $\pm$  2.03                         & 86.11 $\pm$  1.63                     & \textbf{93.48 $\pm$  1.11}              \\
ResNet50      & 49.82 $\pm$  7.29                  & 68.01 $\pm$  6.82                     & 89.14 $\pm$  2.10                         & 81.14 $\pm$  2.14                     & \textbf{92.76 $\pm$  0.86}              \\
MobileNet v2  & 62.47 $\pm$  5.37                  & 77.46 $\pm$  3.77                     & 88.68 $\pm$  0.99                         & 82.51 $\pm$  2.77                     & \textbf{91.79 $\pm$  0.86}              \\
ShuffleNet v2 & 55.02 $\pm$  9.64                  & 73.92 $\pm$  3.01                     & 88.24 $\pm$  1.89                         & 82.97 $\pm$  1.87                     & \textbf{91.31 $\pm$  1.59}              \\
AConvNet      & 78.68 $\pm$  3.79                  & 81.14 $\pm$  3.34                     & 86.92 $\pm$  1.30                         & 80.16 $\pm$  2.88                     & \textbf{87.72 $\pm$  2.29}              \\
AM-CNN        & 82.73 $\pm$  3.72                  & 78.74 $\pm$  6.46                     & 90.62 $\pm$  2.00                         & 87.72 $\pm$  1.79                     & \textbf{93.60 $\pm$  0.73}              \\ \noalign{\global\arrayrulewidth=1.0pt} \hline
\end{tabular}%
}
\end{table*}

To merge the SAMPLE measured data with MSTAR clutter, we used a segmentation algorithm proposed in \cite{ref5}. First, the algorithm is applied to the measured data \( I_{\text{meas}} \) to generate region masks for the target, shadow, and clutter, denoted as $M_{\text{target}}$, $M_{\text{shadow}}$, and $M_{\text{clutter}}$, respectively. The difference between the mean of selected clutter image \( C \) and that of \( I_{\text{meas}} \) in \( M_{\text{clutter}} \) is then added to the target region of \( I_{\text{meas}} \) in \( M_{\text{target}} \), ensuring that the targets maintain a higher intensity than the clutter. After this process, the target and shadow regions of \( I_{\text{meas}} \) are merged into the clutter data to generate the merged measured data \( I_{\text{merged}} \). In this case, \( C \) is randomly selected, except in the shadow region where the target is unlikely to exist. The equations are as follows:
\begin{equation}
    M_{\text{target}}, M_{\text{shadow}}, M_{\text{clutter}} = G(I_{\text{meas}})
\end{equation}
\begin{equation}
    \bar{C} = \frac{\sum_{h,w} C(h, w) \cdot M_{\text{clutter}}(h, w)}{\sum_{h,w} M_{\text{clutter}}(h, w)}
\end{equation}
\begin{equation}
    \bar{I}_{\text{meas}} = \frac{\sum_{h,w} I_{\text{meas}}(h, w) \cdot M_{\text{clutter}}(h, w)}{\sum_{h,w} M_{\text{clutter}}(h, w)}
\end{equation}
\begin{equation}
    d = \bar{C} - \bar{I}_{\text{meas}}
\end{equation}
\begin{equation}
    I_{\text{merged}} = (I_{\text{meas}} + d) \cdot M_{\text{target}} + I_{\text{meas}} \cdot M_{\text{shadow}} + C \cdot M_{\text{clutter}}.
\end{equation}

As shown in Fig.~\ref{fig5}, the GMM can segment the target and clutter regions. However, the reason for incorporating the segmentation algorithm proposed in \cite{ref5} is as follows: although segmentation is crucial in our proposed method, it serves as a foundational element rather than the core of this study. By using a well-established segmentation algorithm specifically designed for SAR data, we reduce the risk of segmentation errors and ensure accuracy and consistency. Further, GMMs struggle to effectively segment shadow regions, whereas the algorithm in \cite{ref5} can accurately segment shadow regions.

The ATR results for each network are summarized in Table~\ref{table3}. Analysis shows that, as in the first scenario, \textbf{SSR} achieved the highest performance across all networks. This demonstrates that the proposed SSR framework enables performance improvements that cannot be achieved by simply adding noise as well as highlights its superiority over basic contrast adjustment methods. Comparing \textbf{Only Noise} and \textbf{SSR w/o Noise}, although the first scenario showed that \textbf{SSR w/o Noise} achieved higher performance across all networks, the second scenario showed that VGG-16 and AConvNet achieved higher performance with \textbf{Only Noise}. This suggests that simply adding noise can be effective in real-world environments in which diverse clutter occurs.

The second scenario provides relevant insights into how models trained on synthetic data perform on real data across various clutter environments. It supports the effectiveness of the proposed SSR method in overcoming the challenges that arise in real operational environments.

\begin{figure*}[t]
\centering
\includegraphics[width=0.9\linewidth]{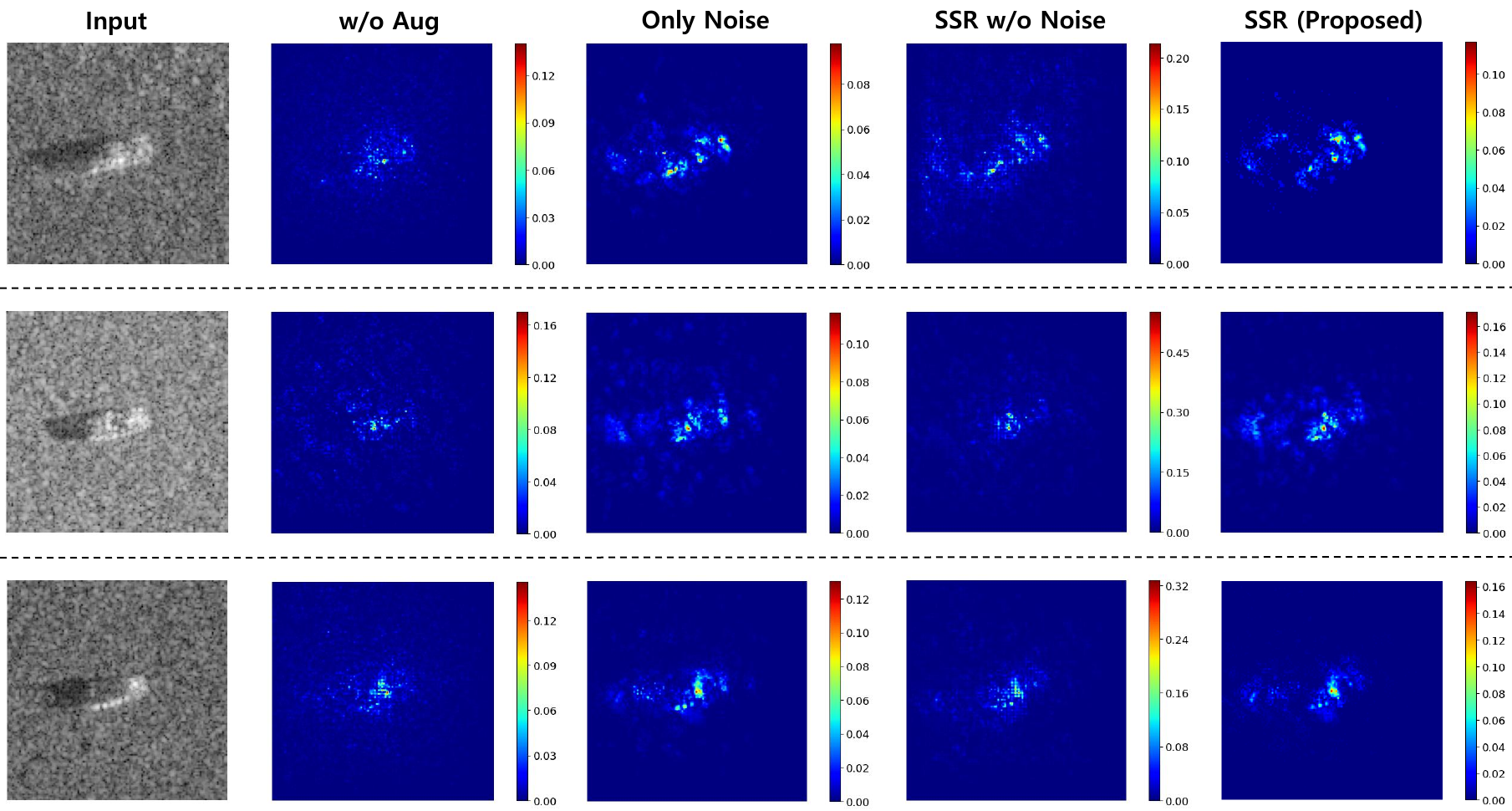}
\caption{2S1, BTR70, M35, in order from above. Using SHAP, the decision point visualization of ResNet18 trained by each method in first scenario. SHAP value is displayed as colorbar on the right side of each result, and is the absolute value used to visualize the area that affects the decision.}
\label{fig9}
\end{figure*}

\subsection{Analysis}
\subsubsection{Decision Points Visualization}
To analyze the effectiveness of the proposed method, we examine which parts of the input data each model relies on for predictions. For this, we employ SHAP (SHapley Additive exPlanations)\cite{ref44} to evaluate the importance of input variables for each model’s prediction. By visualizing the decision points through SHAP analysis, we can gain a concrete understanding of how the proposed augmentation method contributes to target signature recognition. In addition, by comparing the effects of models trained with \textbf{Only Noise}, \textbf{SSR w/o Noise}, and \textbf{SSR}, we can quantitatively explain why SSR is effective.

Fig.~\ref{fig9} shows the SHAP results for the measured data, using ResNet18 trained with each augmentation technique under the first scenario. \textbf{w/o Aug} shows that the models fail to focus adequately on the target region and are heavily influenced by clutter, suggesting that the models do not effectively distinguish between the target and clutter, leading to a higher likelihood of misclassification. By contrast, \textbf{Only Noise} shows that the decision points are more concentrated in the target region, but the SHAP values remain low (refer to the color bar), indicating that the models may be considering irrelevant features and do not focus on the most important domain-agnostic information. \textbf{SSR w/o Noise} exhibits very high SHAP values concentrated on specific scattering topological points, implying that the models identified key domain-agnostic features of the target. However, this also suggests a potential over reliance on these features, which could decrease generalizability. Meanwhile, \textbf{SSR} achieves a balance by having higher SHAP values than \textbf{Only Noise} in specific regions while also being distributed across a wider target region that \textbf{SSR w/o Noise}. This means that \textbf{SSR} combines the ability to consider various features such as shadow and target shapes, which are the advantages of \textbf{Only Noise}, with the ability to focus on certain important domain-agnostic features, which are the advantages of \textbf{SSR w/o Noise}.

As a result, the models can account for various environmental factors while reinforcing in on important features, which can be highly advantageous in SAR-ATR. By reinforcing essential target identification information and minimizing the influence of unnecessary clutter, the proposed method enhances a model’s ability to generalize across different domains.

\begin{figure*}[t]
\centering
\includegraphics[width=0.8\linewidth]{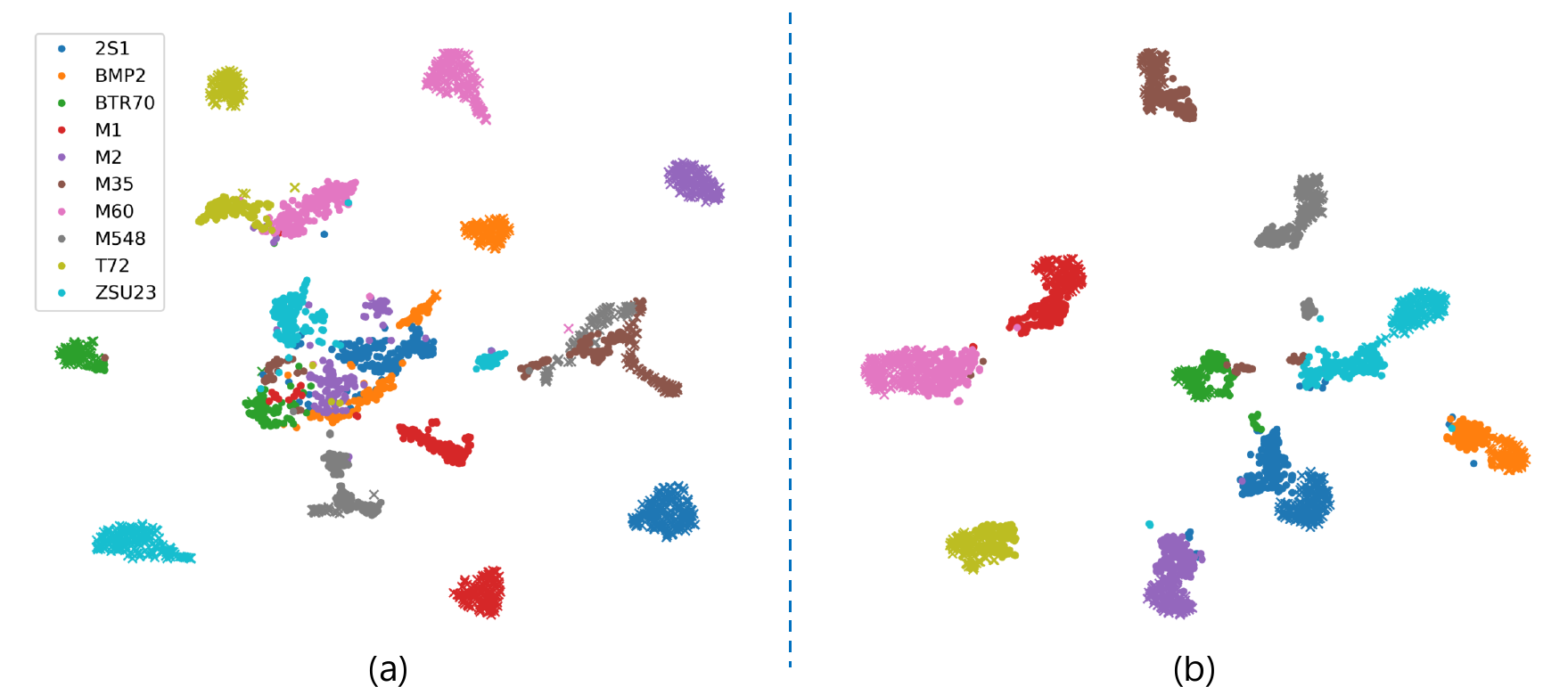}
\caption{Visualization of the feature space using t-SNE from the last convolutional layer of a ResNet18 network trained using SSR. Synthetic data is denoted by $\times$ and measured data is denoted by $\bullet$. The same class is denoted by the same color. (a) Network trained only with synthetic data, (b) Network trained with augmented data using SSR.}
\label{fig10}
\end{figure*}

\subsubsection{Feature Space Visualization}
We visualize the last convolutional layer feature space of ResNet18 trained with \textbf{SSR} in the first scenario using the t-distributed stochastic neighbor embedding (t-SNE) algorithm~\cite{ref51}. This visualization helps further demonstrate the effectiveness of SSR by showing how a model trained on only synthetic data, also distinguishes between classes in the measured data.

t-SNE is an excellent tool for reducing the complex structure of high-dimensional data to lower dimensions for visualization, clearly showing the separation between classes. Specifically, it visually represents the relationship between synthetic and measured data, highlighting the effectiveness of SSR in resolving domain discrepancy issues. 

As shown in Fig.~\ref{fig10}, the t-SNE-transformed feature space shows that, except for a few misclassified data points, the measured data are clearly separated by class, indicating that SSR enables high performance on measured data without prior knowledge of the target domain. This improvement in feature distribution supports the consistency of the proposed method across various classes.

\section{Ablation Study}

\subsection{Comparison of Soft and Hard Segmentation in Target and Clutter Regions}
In the ablation study, we compare the effectiveness of the proposed SSR framework with an alternative approach that uses a hard segmentation algorithm to completely separate the target and clutter regions. The hard segmentation approach is tested in two ways. 1) Gaussian noise is added to the entire image, and the clutter intensity is modified. This not only introduces changes across the entire image but also results in additional modifications to the mean clutter intensity. 2) Gaussian noise is added, and the intensity is modified only in the clutter region, leaving the target region unchanged. This focuses the changes exclusively on the clutter region. These two approaches are compared with the SSR framework under the same experimental setup as in the first scenario.

The ATR results for each network are summarized in Table~\ref{table4}. When comparing the hard segmentation approaches with \textbf{SSR}, \textbf{SSR} achieved higher performance across all networks. Although the hard segmentation approaches clearly separate the target and clutter regions, they can introduce boundary artifacts that do not naturally occur in real-world scenarios. These artifacts, particularly at the boundaries between regions, can disrupt the natural transition between the target and clutter, causing the network to fail in accounting for the complexity of real-world environments. As a result, the network may fail to adequately capture shape information in real-world data, leading to reduced generalizability.

\begin{table*}[]
\centering
\caption{Results of ATR accuracy [\%] using different segmentation methods}
\label{table4}
\resizebox{0.53\textwidth}{!}{%
\begin{tabular}{lccccc}
\noalign{\global\arrayrulewidth=1.0pt}
\hline
\noalign{\global\arrayrulewidth=0.4pt}
\multicolumn{1}{c}{} & \textbf{1) Hard Seg}    & \textbf{2) Hard Seg}    & \textbf{SSR (Proposed)} \\ \hline
VGG-11               & 78.53 $\pm$ 1.40 & 67.35 $\pm$ 2.68 & \textbf{87.51 $\pm$ 0.89}     \\
VGG-16               & 74.71 $\pm$ 1.90 & 68.29 $\pm$ 2.96 & \textbf{87.64 $\pm$ 1.52}     \\
ResNet18             & 74.71 $\pm$ 1.99 & 77.61 $\pm$ 6.39 & \textbf{94.69 $\pm$ 1.19}     \\
ResNet50             & 73.05 $\pm$ 2.05 & 70.48 $\pm$ 9.45 & \textbf{92.95 $\pm$ 2.22}     \\
MobileNet v2         & 76.42 $\pm$ 2.09 & 68.87 $\pm$ 9.18 & \textbf{91.97 $\pm$ 1.98}     \\
ShuffleNet v2        & 74.10 $\pm$ 1.69 & 72.37 $\pm$ 5.78 & \textbf{91.95 $\pm$ 1.30}     \\
AConvNet             & 78.71 $\pm$ 3.57 & 82.24 $\pm$ 4.65 & \textbf{91.15 $\pm$ 2.59}     \\
AM-CNN               & 75.79 $\pm$ 1.90 & 87.88 $\pm$ 2.67 & \textbf{94.38 $\pm$ 1.20}     \\ \noalign{\global\arrayrulewidth=1.0pt}\hline
\end{tabular}}
\end{table*}

\begin{figure}[ht]
    \centering
    \begin{minipage}[t]{0.4\textwidth}
        \centering
        \includegraphics[width=\textwidth]{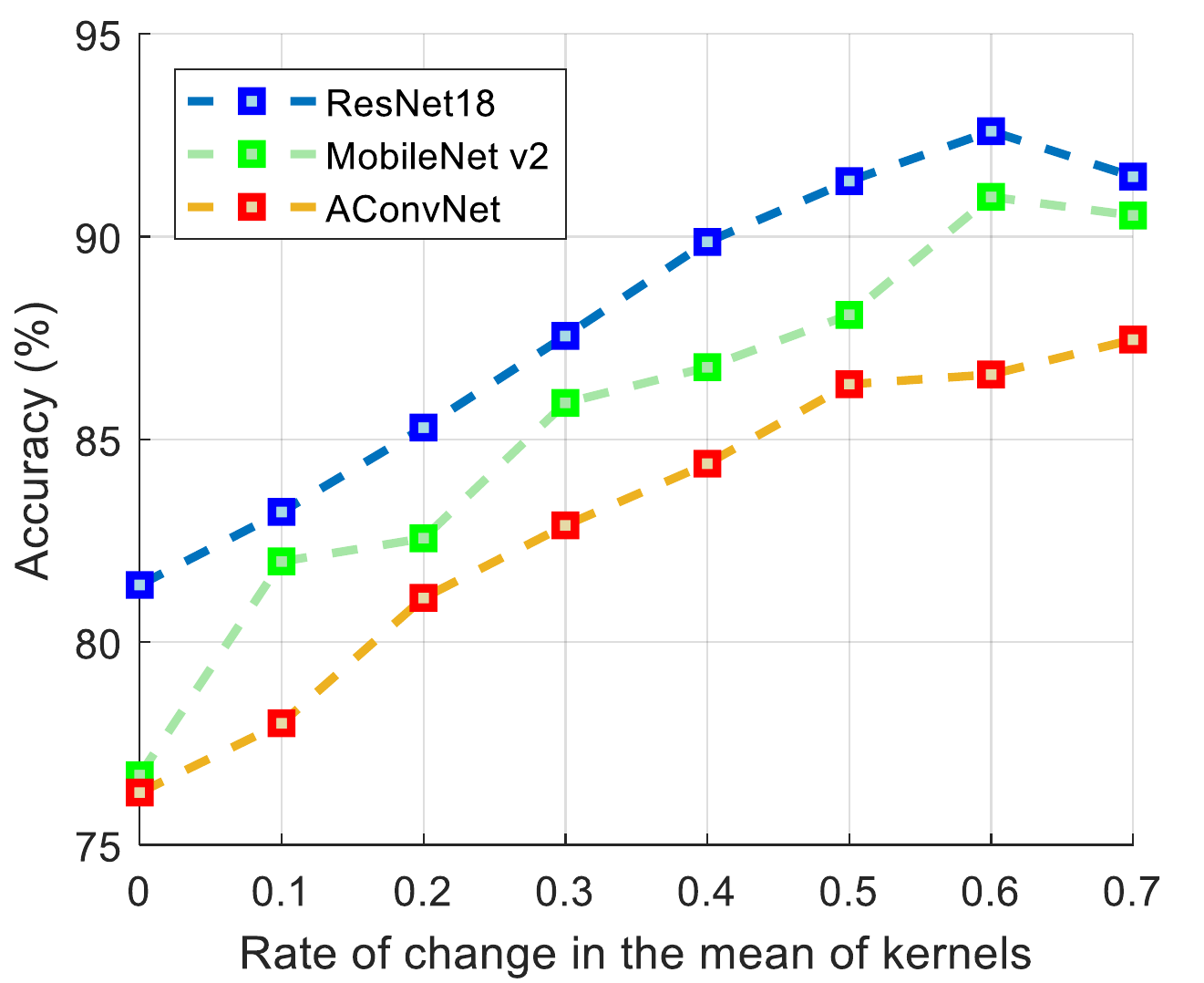} 
        \caption{Effect of kernel mean change rate on network accuracy across different models}
        \label{fig12}
    \end{minipage}%
    \hspace{0.05\textwidth} 
    \begin{minipage}[t]{0.4\textwidth}
        \centering
        \includegraphics[width=\textwidth]{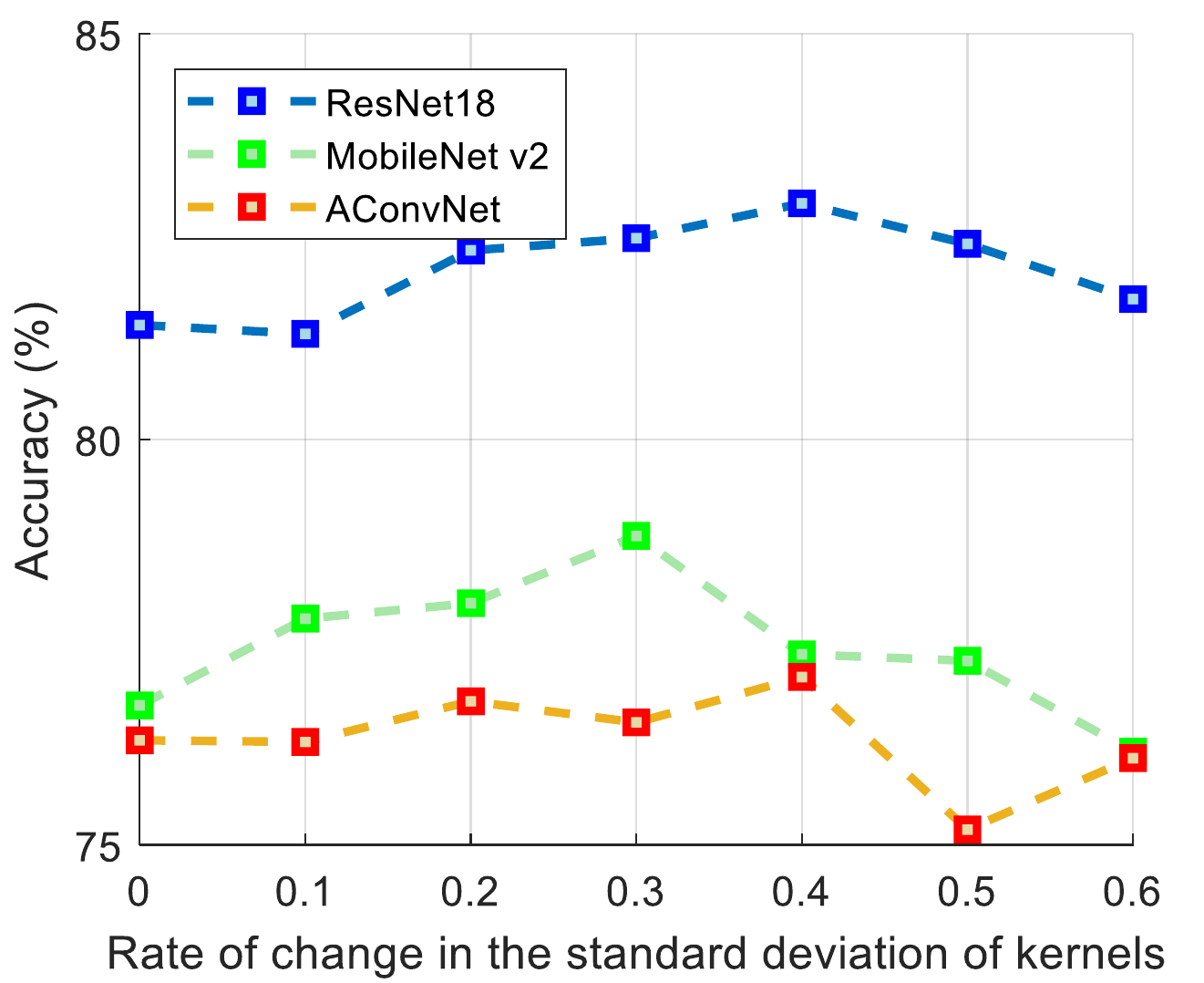} 
        \caption{Effect of kernel standard deviation change rate on network accuracy across different models}
        \label{fig14}
    \end{minipage}
\end{figure}

\subsection{Hyperparameter Selection}
The effective application of the proposed SSR framework relies heavily on the selection of the Gaussian kernel's mean change rate, $\alpha$, and standard deviation change rate, $\beta$. These two hyperparameters control intensity adjustments in the clutter region and ultimately have a significant impact on network performance. We conducted experiments with various $\alpha$ and $\beta$ values; the results are shown in Fig.~\ref{fig12} and \ref{fig14}. 

Fig.~\ref{fig12} shows the changes in accuracy with varying $\alpha$. We observed that accuracy increased with higher $\alpha$ in ResNet18, MobileNet v2, and AConvNet, with the optimal performance observed at $\alpha = 0.6$ for ResNet18 and MobileNet v2. As shown in Fig.~\ref{fig14}, the impact of $\beta$ on performance was less pronounced; however, there was a slight performance improvement across all networks when $\beta$ was set between 0.2 and 0.4 compared with when it was set to 0 indicating that $\beta$ also contributes to addressing domain discrepancy issues. Therefore, we concluded that setting $\alpha$ to 0.6 and $\beta$ to 0.4 is most appropriate for achieving optimal performance.

\subsection{Measured Data Augmentation}
In the ablation study, we apply the SSR framework to the MSTAR dataset, focusing on the measured data. Owing to the minimal difference in speckle noise between the training and testing data, a weaker Gaussian noise with a standard deviation of 0.2 was used while the other parameters are as in the original SSR experiments. The experiments are structured into two scenarios.

\subsubsection{First Scenario}
In this scenario, data with a depression angle of 17$^\circ$ across 10 classes were used for training, with each class having 36 samples varied by approximately 10$^\circ$ of the aspect angle. To evaluate model generalizability, testing was performed using data with a depression angle of 15$^\circ$ (Table~\ref{table5}). The purpose of this experiment is to assess whether applying SSR improves the generalizability of models trained on a limited amount of measured data.

As shown in Table~\ref{table6}, applying SSR led to performance improvements across all networks. Notably, significant enhancements were observed in MobileNet v2 and ShuffleNet v2, indicating that SSR helps models better generalize across various angles and conditions even with a limited amount of measured data.

\begin{table*}[]
\centering
\caption{Training and testing datasets under experimental scenarios with measured data}
\label{table5}
\resizebox{0.65\textwidth}{!}{%
\begin{tabular}{cccccc}
\noalign{\global\arrayrulewidth=1.0pt}
\hline
\noalign{\global\arrayrulewidth=0.4pt}
\multirow{2}{*}{\textbf{Class}} & \multirow{2}{*}{\textbf{Serial No.}} & \multicolumn{2}{c}{\textbf{First Scenario}} & \multicolumn{2}{c}{\textbf{Second Scenario}}     \\ \cline{3-6} 
                                &                                      & Train (17$^\circ$)        & Test (15$^\circ$)       & Train (17$^\circ$) & Test (30$^\circ$) + Clutter \\ \hline
2S1                             & B01                                  & 36                        & 274             & 299                & 288                         \\
BMP2                            & 9563                                 & 36                        & 195             & -                  & -                           \\
BRDM2                           & E71                                  & 36                        & 274             & 298                & 287                         \\
BTR60                           & 7532                                 & 36                        & 195             & -                  & -                           \\
BTR70                           & C71                                  & 36                        & 196             & -                  & -                           \\
D7                              & 13015                                & 36                        & 274             & -                  & -                           \\
T62                             & A51                                  & 36                        & 273             & -                  & -                           \\
T72                             & 132                                  & 36                        & 196             & 232                & -                           \\
                                & 812                                  & -                         & -               & 231                & -                           \\
                                & S7                                   & -                         & -               & 228                & -                           \\
                                & A64                                  & -                         & -               & -                  & 288                         \\
ZIL131                          & E12                                  & 36                        & 274             & -                  & -                           \\
ZSU23-4                         & D08                                  & 36                        & 274             & 299                & 288                         \\ \noalign{\global\arrayrulewidth=1.0pt} \hline
\end{tabular}
}
\end{table*}

\begin{table}[ht]
\centering
\begin{minipage}[t]{0.4\textwidth}
    \centering
    \caption{Results of the first scenario experiments with measured data ATR accuracy [\%]}
    \label{table6}
    \resizebox{\textwidth}{!}{%
    \begin{tabular}{lccccc}
    \noalign{\global\arrayrulewidth=1.0pt}
    \hline
    \noalign{\global\arrayrulewidth=0.4pt}
    \multicolumn{1}{c}{} & \textbf{w/o Aug}    & \textbf{SSR (Proposed)} \\ \hline
    VGG-11               & 83.05 $\pm$ 1.10 & \textbf{87.15 $\pm$ 1.44}     \\
    VGG-16               & 76.29 $\pm$ 2.28 & \textbf{83.67 $\pm$ 2.05}     \\
    ResNet18             & 81.48 $\pm$ 3.37 & \textbf{92.81 $\pm$ 0.94}     \\
    ResNet50             & 69.74 $\pm$ 3.37 & \textbf{90.86 $\pm$ 1.31}     \\
    MobileNet v2         & 42.52 $\pm$ 3.77 & \textbf{91.53 $\pm$ 1.94}     \\
    ShuffleNet v2        & 58.00 $\pm$ 5.23 & \textbf{90.68 $\pm$ 1.54}     \\
    AConvNet             & 81.79 $\pm$ 2.06 & \textbf{82.47 $\pm$ 1.96}     \\
    AM-CNN               & 86.48 $\pm$ 5.43 & \textbf{96.08 $\pm$ 0.56}     \\ \noalign{\global\arrayrulewidth=1.0pt}\hline
    \end{tabular}}
\end{minipage}%
\hspace{0.05\textwidth} 
\begin{minipage}[t]{0.4\textwidth}
    \centering
    \caption{Results of the second scenario experiments with measured data ATR accuracy [\%]}
    \label{table7}
    \resizebox{\textwidth}{!}{%
    \begin{tabular}{lccccc}
    \noalign{\global\arrayrulewidth=1.0pt}
    \hline
    \noalign{\global\arrayrulewidth=0.4pt}
    \multicolumn{1}{c}{} & \textbf{w/o Aug}    & \textbf{SSR (Proposed)} \\ \hline
    VGG-11               & 56.27 $\pm$ 3.93 & \textbf{86.66 $\pm$ 2.15}     \\
    VGG-16               & 47.49 $\pm$ 8.93 & \textbf{79.05 $\pm$ 4.01}     \\
    ResNet18             & 72.61 $\pm$ 1.46 & \textbf{81.61 $\pm$ 2.20}     \\
    ResNet50             & 71.35 $\pm$ 5.20 & \textbf{80.44 $\pm$ 3.61}     \\
    MobileNet v2         & 63.94 $\pm$ 6.36 & \textbf{86.75 $\pm$ 2.76}     \\
    ShuffleNet v2        & 65.69 $\pm$ 3.02 & \textbf{82.80 $\pm$ 3.10}     \\
    AConvNet             & 77.45 $\pm$ 2.33 & \textbf{88.68 $\pm$ 2.33}     \\
    AM-CNN               & 75.26 $\pm$ 5.66 & \textbf{86.68 $\pm$ 4.23}     \\ \noalign{\global\arrayrulewidth=1.0pt}\hline
    \end{tabular}}
\end{minipage}
\end{table}

\subsubsection{Second Scenario}
This scenario simulates an extended operating condition (EOC) by training models on data with a depression angle of 17$^\circ$ across four classes and testing them on data with a significantly different depression angle of 30$^\circ$ (Table~\ref{table5}). In addition, for the T72 class, the serial numbers of the data used for training and testing differ. This experiment assesses how effectively SSR can minimize performance degradation caused by the angle and serial number differences.

According to the results in Table~\ref{table7}, SSR applied models demonstrated strong adaptability to the differences in angle between training and testing data, even under EOC. Changes in the depression angle and serial numbers significantly alter the target signatures. Nevertheless, SSR-applied models can overcome these discrepancies, thereby mitigating performance degradation and maintaining higher accuracy. This demonstrates that SSR is effective even under EOC and is useful for reducing domain discrepancies under various operational conditions.

\section{Conclusion}
In this study, we proposed a novel data augmentation framework, SSR, to mitigate domain discrepancies between synthetic and measured SAR data in ATR tasks. By softly segmenting and randomizing the clutter and target regions in the synthetic data using a GMM, SSR effectively reduces differences in the statistical properties of the synthetic and measured data, thereby improving the generalizability of deep learning models. Overall, this study addresses the limitations of synthetic SAR data, making them more applicable to real-world scenarios in which data acquisition is challenging or impossible. Future research could explore the integration of SSR with other DA or DG techniques to further enhance its effectiveness.

\bibliographystyle{unsrt}
\bibliography{ref}

\end{document}